\documentclass[journal,twoside]{IEEEtran}
\ifCLASSINFOpdf
\else
\fi

\usepackage{graphics} 
\usepackage{graphicx}
\usepackage[caption=false,font=footnotesize]{subfig}
\usepackage{eufrak}
\usepackage{epsfig} 
\usepackage{amsmath} 
\usepackage{amsfonts}
\usepackage{amssymb}  
\usepackage{blindtext}
\usepackage{siunitx}
\usepackage{nicefrac}
\usepackage{cite}
\usepackage{xcolor}
\usepackage{multirow}
\usepackage{multicol}
\usepackage[%
  xindy,
  section = section,
  nonumberlist,
  sanitize={symbol=false},
  shortcuts,
  acronym,
  nomain,
  nowarn
]{glossaries}                           

\newacronym{rl}{RL}{Reinforcement Learning}
\newacronym{ml}{ML}{Machine Learning}
\newacronym{mlp}{MLP}{Multi Layer Perceptron}
\newacronym{bo}{BO}{Bayesian Optimization}
\newacronym{mpc}{MPC}{model predictive control}
\newacronym{wlog}{w.l.o.g.}{without loss of generality}
\newacronym{LF}{LF}{left front}
\newacronym{RF}{RF}{right front}
\newacronym{LH}{LH}{left hind}
\newacronym{RH}{RH}{right hind}
\newacronym{wrt}{w.r.t.}{with respect to}
\newacronym{subt}{SubT}{DARPA Subterranean Challenge}
\newacronym{gp}{GP}{Gaussian Process}
\newacronym{hebo}{HEBO}{Heteroscedastic Evolutionary Bayesian Optimisation}
\newacronym{nn}{NN}{Neural Network}
\newacronym{cot}{CoT}{Cost of Transportation}
\newacronym{cotr}{CoTr}{Cost of Torque}
\newacronym{sea}{SEA}{Series Elastic Actuator}
\newacronym{pea}{PEA}{Parallel Elastic Actuator}
\newacronym{ai}{AI}{Artificial Intelligence}
\newacronym{dof}{DoF}{Degrees of Freedom}
\newacronym{haa}{HAA}{hip abduction / adduction}
\newacronym{hfe}{HFE}{hip flexion / extension}
\newacronym{kfe}{KFE}{knee flexion / extension}
\newacronym{aops}{AoPS}{ANYmal on Parallel Springs}
\newacronym{ftg}{FTG}{Foot Trajectory Generator}
\newacronym{ik}{IK}{Inverse Kinematics}
\newacronym{gru}{GRU}{Gated Recurrent Unit}
\newacronym{rnn}{RNN}{Recurrent Neural Network}
\newacronym{ppo}{PPO}{Proximal Policy Optimization}
\newacronym{mdp}{MDP}{Markov Decision Process}
\newacronym{tcn}{TCN}{Temporal Convolutional Network}

\makeglossaries

\DeclareMathOperator{\argmin}{arg\,min}

\newcommand*{\tran}{^\mathsf{T}}

\renewcommand{\baselinestretch}{0.925}

\begin{document}
\bstctlcite{IEEEexample:BSTcontrol}
%
\title{Learning-based Design and Control for Quadrupedal Robots with Parallel-Elastic Actuators}
%
%
%

\author{Filip Bjelonic$^{1,2}$, Joonho Lee$^{1}$, Philip Arm$^{1}$, Dhionis Sako$^{1}$, Davide Tateo$^{2}$, Jan Peters$^{2}$, Marco Hutter$^{1}$
\thanks{Manuscript received: August 27, 2022; Revised November 21, 2022; Accepted December 16, 2022. }
\thanks{This paper was recommended for publication by Editor Abderrahmane A. Kheddar upon evaluation of the Associate Editor and Reviewers’ comments. 
This work was supported by a fellowship within the IFI program of the German Academic Exchange Service (DAAD).}
\thanks{$^{1}$ Authors are with ETH Zurich; Robotic Systems Lab; Leonhardstrasse 21, 8092 Zurich, Switzerland. 

$^{2}$ Authors are with TU Darmstadt; Intelligent Autonomous Systems Lab; Hochschulstrasse 10, 64289 Darmstadt, Germany}%
\thanks{Digital Object Identifier (DOI): see top of this page.}
}

%
%

\markboth{IEEE Robotics and Automation Letters, Preprint Version. Accepted December, 2022}%
{Bjelonic \MakeLowercase{\textit{et al.}}: Learning-based design and control for quadrupedal robots}
%



\maketitle


\begin{abstract}
Parallel-elastic joints can improve the efficiency and strength of robots by assisting the actuators with additional torques. For these benefits to be realized, a spring needs to be carefully designed. However, designing robots is an iterative and tedious process, often relying on intuition and heuristics. 
We introduce a design optimization framework that allows us to co-optimize a parallel elastic knee joint and locomotion controller for quadrupedal robots with minimal human intuition.
We design a parallel elastic joint and optimize its parameters with respect to the efficiency in a model-free fashion.
In the first step, we train a design-conditioned policy using model-free Reinforcement Learning, capable of controlling the quadruped in the predefined range of design parameters. Afterwards, we use Bayesian Optimization to find the best design using the policy. We use this framework to optimize the parallel-elastic spring parameters for the knee of our quadrupedal robot ANYmal together with the optimal controller. We evaluate the optimized design and controller in real-world experiments over various terrains. Our results show that the new system improves the torque-square efficiency of the robot by \SI{33}{\percent} compared to the baseline and reduces maximum joint torque by \SI{30}{\percent} without compromising tracking performance. The improved design resulted in \SI{11}{\percent} longer operation time on flat terrain.
\end{abstract}



%
\IEEEpeerreviewmaketitle

\section{Introduction}
\IEEEPARstart{T}{he} quest of creating a single versatile, efficient and strong robotic platform has driven research in legged robotics for many years. While controllers are getting more robust and intelligent, locomotion performance is limited by the available joint speed and joint torque. Better performance can be achieved by creating more efficient and powerful actuators. Adding elastic elements has the promise of supporting the actuators with additional torque \cite{abate2018mechanical}.

\begin{figure}
    \centering
    \includegraphics[width=0.489\textwidth]{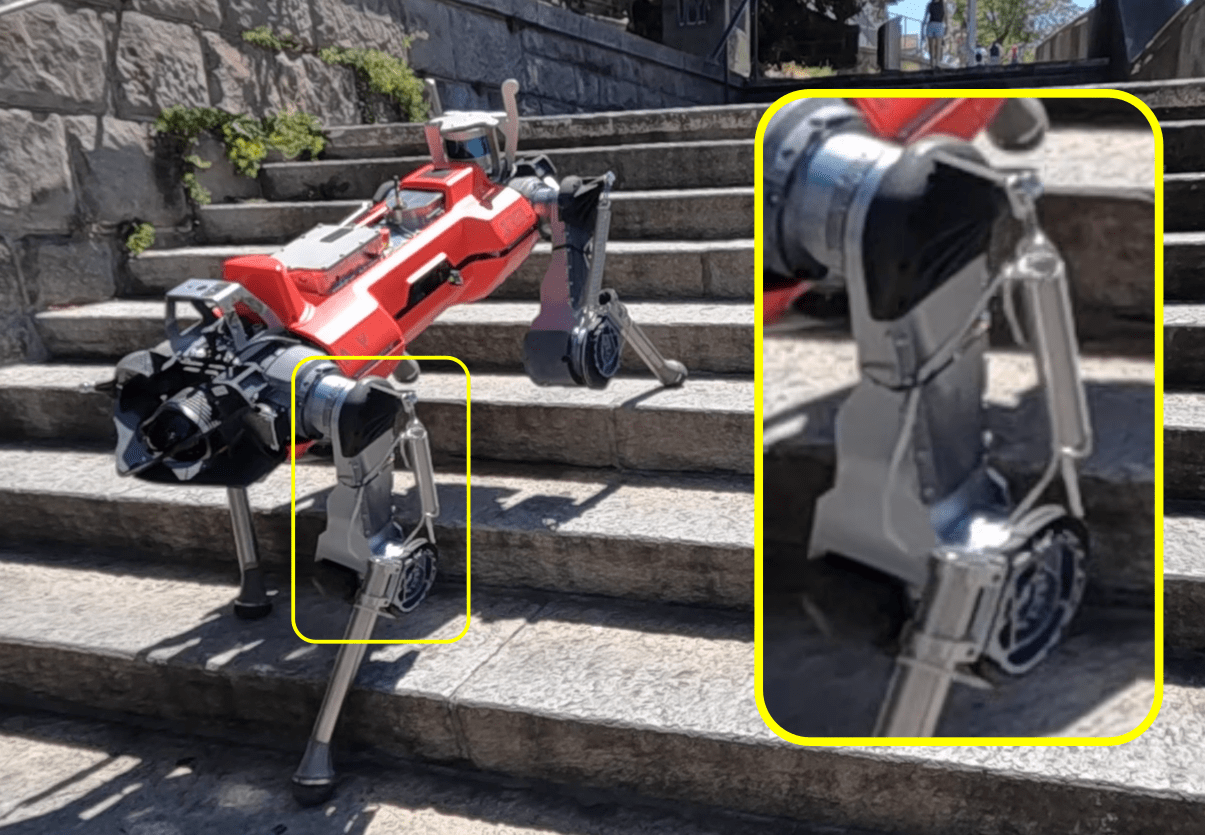}
    \caption{The ANYmal robot with parallel-elastically actuated knee joints. ANYmal is walking upstairs at the central station of Zurich, which is used as one of the experimental sites during this work.}
    \label{fig:aops_teaser}
\end{figure}

In this letter, we explore the effect of the elastic component on energy efficiency during locomotion by attaching a parallel spring mechanism on the knee joints of the ANYmal robot (Fig.~\ref{fig:aops_teaser}).
This system is used to experiment and verify the benefit of the parallel elasticity.

\subsection{Robots with elastic actuators}

One of the first approaches in this direction was the \ac{sea} by Gill Pratt \cite{pratt1995series} which incorporates a series-elastic element between the actuator and the load. This design makes the joint positioning error-tolerant, reduces impact loads, and, most importantly, allows for precise torque measurement. The ANYmal quadrupedal robot \cite{hutter2016anymal} integrates into its ANYdrive actuator a serial elastic spring. More examples are ATRIAS \cite{hubicki2016atrias}, a biped that has serial elastic springs at the actuator level and Cassie \cite{abate2018mechanical} with a 6-bar linkage with 2 springs in series. HyQ \cite{semini2011design} has a serial elastic spring between the knee and the foot of the robot, which reduces foot chattering during touch-down.

Another approach is the \ac{pea}. In this setup, the actuator and the spring are in parallel. While this approach has been studied in robotic manipulation for gravity compensation \cite{kashiri2018overview}, for pick-and-place \cite{scalera2019energy} and efficient oscillation \cite{bjelonic2022experimental}, there is no comparative evaluation of walking robots with \ac{pea} outside of controlled lab environments. One example of a legged robot with \ac{pea}s is SpaceBok \cite{arm2019spacebok}. In a lab experiment with simulated moon gravity, \ac{pea}s reduced the energy required for a jump by a factor of two on this robot \cite{kolvenbach2019towards}. Another, more recent example of using \ac{pea} is BirdBot \cite{badri2022birdbot}, which has a parallel elastic spring clutching mechanism, spanning multiple joints. The avian-inspired leg design shows self-stable and robust bipedal locomotion while requiring \SI{10}{\percent} of the knee-flexing torque compared to a non-clutching parallel spring setup. Another example is STEPPR \cite{mazumdar2016parallel}. This bipedal robot has a parallel-elastic spring at the hip and the ankle. Using only the hip springs during walking, the robot consumes \SI{31}{\percent} less joint electrical power and reduces power consumption overall by \SI{13}{\percent}.

All of the previous works mention the possibility of saving energy with the carefully designed springs. Unfortunately, most of them are designed based on heuristic and cannot exploit the full potential of elastic elements. 


Building upon intuitive design, a common approach starts with mimicking nature's counterparts \cite{silva2012literature}. Atrias \cite{hubicki2016atrias} and BirdBot \cite{badri2022birdbot} for instance are inspired by ostriches and the emu. The problem with bio-inspired design is the high amount of variables that need to be taken into account to fully model the targeted animal accurately. Nevertheless, there is no systematic way of designing robots in general.

\subsection{Computational design}\label{sec:robot_design}

Computational robot design can be divided into gradient-based methods that work well with deterministic differentiable objective functions, gradient-free algorithms with smooth objectives (e.g. trust-region methods), meta-heuristic methods that are nature inspired (e.g. simulated annealing, genetic algorithms), and surrogate methods (e.g. \ac{bo}) \cite{koziel2011computational}. Meta-heuristic and surrogate methods have been successfully used in black-box optimization problems, where the properties of the objective function are not known in advance \cite{turner2021bayesian} \cite{frazier2018tutorial}.


A work related to the goal in this work has been done by Scalera et al. \cite{scalera2019energy} where the design optimization of elastic elements was carried out for a four \ac{dof} parallel robotic arm. Here, the robot achieved an efficiency gain of \SI{67}{\percent} on a predefined trajectory by defining a non-linear optimization problem for finding energy optimal spring parameters. This approach is unsuitable for legged locomotion since it optimizes over a fixed trajectory that is by no means guaranteed to be optimal.

The approach from De Vincenti et. al. \cite{de2021control} uses a differentiable trajectory tracking controller such that the overall leg design optimization becomes control-aware. Effectively, the gradient computation takes the control formulation into account in each step. Nevertheless, the trajectory is still fixed for all the tasks.

A co-optimization approach is developed by Dinev et. al. \cite{dinev2022versatile} for leg lengths, joint positions, trunk shape, and weight distribution. Here, motion planning is recomputed in every evaluation of the design process. Using finite differences, the design optimization increases the energy efficiency of the Solo robot by a factor of 3 and shows faster convergence than using an evolutionary optimizer (CMA-ES). Using finite differences on rough terrain may result in an unstable solver, making this approach hard to incorporate into our goals.

In general, these methods incorporate a design optimization that is wrapped around the robot control and planning loop. While the approaches incorporate gradient-based or gradient-free solvers for the outer loop, the inner loop can be either fixed \cite{scalera2019energy} \cite{chadwick2020vitruvio}, or efficiently re-optimized in every performance evaluation \cite{dinev2022versatile} \cite{de2021control} \cite{zhao2020robogrammar} \cite{ha2017joint}.

An interesting simultaneous approach from Chen et al. \cite{pmlr-v155-chen21a} defines a hardware policy besides the control policy, that is jointly optimized over the training process with model-free \ac{rl}. The optimized weights of the hardware policy define the hardware parameters and, together with the control policy, create the output of the algorithm. While this is a fully integrated approach, defining the hardware policy as a computational graph is not possible in many cases \cite{pmlr-v155-chen21a}.

Another method by Schaff et al. \cite{8793537} optimizes an \ac{rl} policy and distribution of design parameters at the same time. The agent is able to observe design parameters while the design space slowly shrinks toward high-performing designs. This approach has been successfully applied on a soft robot crawler \cite{schaff2022soft} and outperformed a baseline design from an expert with the optimal design walking more than $2\times$ as fast.

Inspired by the co-optimization approach from Dinev et. al. \cite{dinev2022versatile} and the learning-based approach by Chen et al. \cite{pmlr-v155-chen21a}, the following section briefly introduces our design optimization framework as well as our main contributions.

\subsection{Contribution}

We present a systematic approach to designing elastic mechanisms for legged robots by incorporating design-conditioned controllers in the optimization. 
In particular, we present:
\begin{itemize}
    \item Co-optimization of the design parameters and the locomotion controller for the \ac{pea}-driven legged robot using model-free \ac{rl} and \ac{bo}.
    \item Integration of the optimized design onto the physical system and sim-to-real transfer of the learned control policy.
    \item Real-world experiments to demonstrate the feasibility and robustness of our approach followed by the quantitative evaluation.
\end{itemize}
We would like to emphasize the last contribution because, to the authors' knowledge, this paper provides the first evaluation of \ac{pea}s on walking robots outside of lab environments.
\section{Method}\label{sec:method}

In this section, we first present our \ac{pea} design and then present our framework to co-optimize the controller as well as design parameters. For any equation, vectors and matrices are marked in bold text. Further, we refer to specific legs by their position with respect to the base in the anterior and lateral direction with the \ac{LF}, \ac{RF}, \ac{LH}, and \ac{RH} leg.

\subsection{Parallel Elastic Knee}
\begin{figure}
        \centering
          \hspace*{\fill}%
        \subfloat[Design
        \label{fig:pea}]{
          \includegraphics[height=0.248\textwidth]{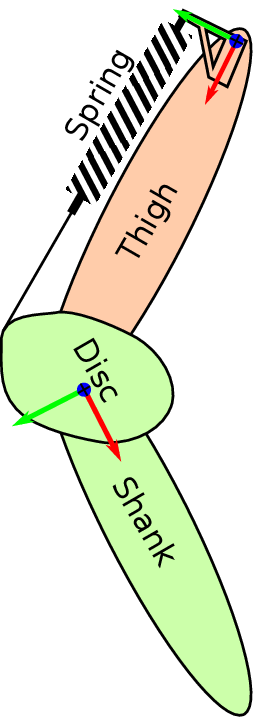}
        }
        \hfill
        \subfloat[Elliptic Cam
        \label{fig:elliptic}]{
          \includegraphics[width=.2\textwidth]{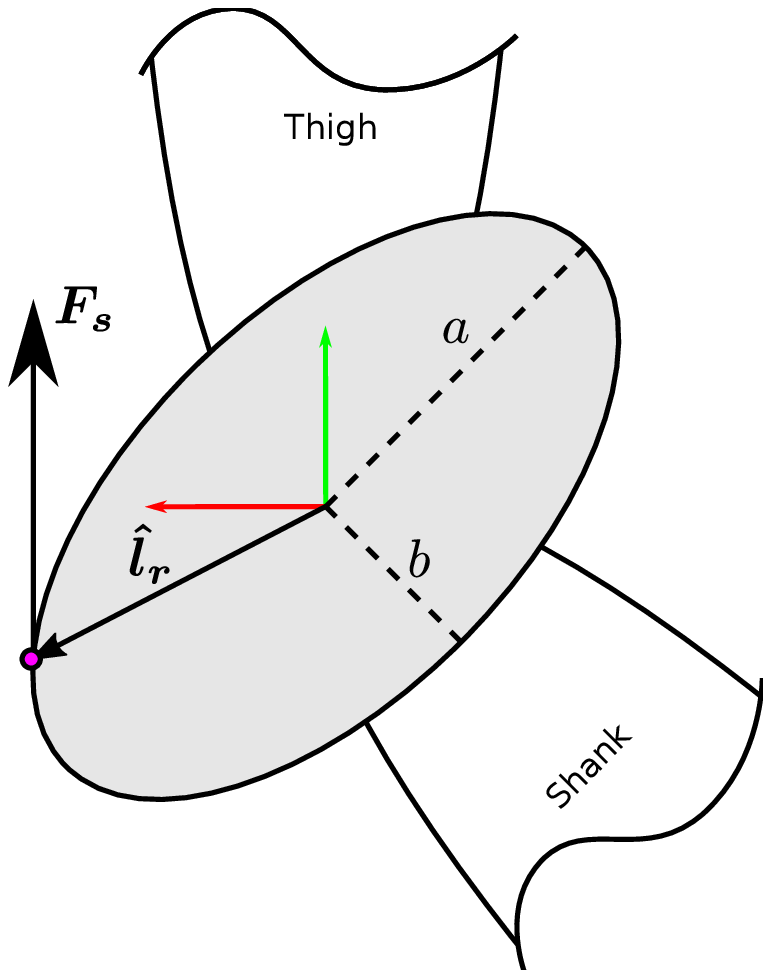}
        }
          \hspace*{\fill}%
    \caption{Fig. \ref{fig:pea} illustrates a generic two-segment leg with potentially nonlinear parallel elastic knee joints. The conceptual implementation of the rotatory spring stiffness $k$ in this work is visualized in Fig. \ref{fig:pea}. The linear elastic spring-wire mechanism connects the thigh with the shank. This creates a spring torque $\tau_s$ on the knee. Parts with the same color are physically connected.}
    \label{fig:pea_concept}
\end{figure}

We design a \ac{pea} knee joint for quadrupedal robots seen in Fig. \ref{fig:pea}. Particularly, a parameterization $\boldsymbol{d} \in \mathcal{D}$ of the joint stiffness $k$ is necessary. We design and implement a spring-wire mechanism (Fig.~\ref{fig:pea}). The wire connects the thigh and shank over a generic disc that defines the lever arm for the spring force. The disc is attached to the shank of the robot.

The torque on the knee that is generated by this design can be calculated in general by

\begin{equation} \label{eq:chap3:spring_torque_generic}
    \boldsymbol{\tau_s}(q) = \boldsymbol{F_s} \times \boldsymbol{\hat{l}_r}(\theta)
\end{equation}
with $\boldsymbol{F_s}$ being the force created by the linear spring, $\theta \in [0, 2\pi)$ defines the boundary of the cam and $\boldsymbol{\hat{l}_r}(\theta)$ is the spring force's lever arm. The amplitude of the spring force can be calculated by Hooke's law as $f_s = ||\boldsymbol{F_s}|| = k_s \Delta l_s$, with $k_s$ being the spring stiffness. The spring elongation $\Delta l_s$ is influenced by the length of wire which is wrapped around the cam and the position of the lever arm. With this setup, the first parameter $d_1$ is the equilibrium position $\bar{q}_{\mathrm{KFE}}$ of the linear spring, which is defined as the knee angle where $\boldsymbol{F_s} = \boldsymbol{0}$. Further parameters are added through the definition of the cam. 
Since the wire is always assumed to be in contact with the cam, the lever arm can be calculated by finding the point on the cam that is tangent to the spring force. This can be formalized by the following equation

\begin{equation} \label{eq:chap3:find_point}
    \boldsymbol{0} = \boldsymbol{F_s} \times \frac{\partial \boldsymbol{\hat{l}_r}}{\partial \theta}.
\end{equation}

We select an elliptic cam as a trade-off between simplicity and degrees of freedom of parameterization. In this case, this equation has always two solutions depending on the side at which the spring force acts. In our case, the left side of the lever arm respects the inequality

\begin{equation} \label{eq:chap3:correct_theta_star}
    \left <\boldsymbol{F_s}, \boldsymbol{\hat{l}_r}(\theta) \right > \geq 0.
\end{equation}

Following, we describe the elliptic cam attached to the shank of the robot.

\subsubsection{Elliptic Cam}\label{sec:elliptic_shape}
\begin{figure}
    \centering
    \includegraphics[width=.3\textwidth]{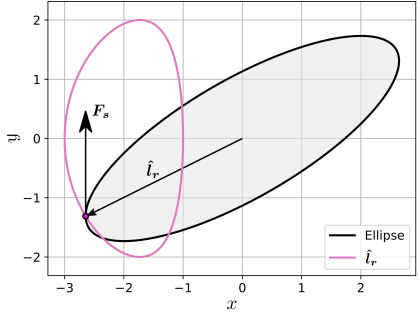}
    \caption{The non-linear trajectory of the spring force's lever arm $\boldsymbol{\hat{l}_r}$ over on full rotation is plotted in pink. The radius of the major and minor axis is $3$ and $1$ respectively, while the spring force is assumed to always point upwards. The length, as well as the angle of the lever arm, changes dynamically, depending on the angle of the knee.}
    \label{fig:lever_arm}
\end{figure}

Elliptic cam is defined by
\begin{align}
    \boldsymbol{l_r}(\theta) &=
    \boldsymbol{R}_\phi
    \begin{bmatrix}
    a \cdot cos(\theta) \\
    b \cdot sin(\theta)
    \end{bmatrix} \label{eq:chap3:parameter_ellipse}\\
    \boldsymbol{R}_\phi &=     \begin{bmatrix}
    cos(\phi) & -sin(\phi) \\
    sin(\phi) & cos(\phi)
    \end{bmatrix} \nonumber \\
    \phi &= \phi_0 + q_\mathrm{KFE} \nonumber,
\end{align}
where $\theta \in [0, 2\pi)$ and $\phi_0$ being the initial angle of the ellipse with respect to the shank's longitudinal axis at $q_\mathrm{KFE} = \SI{0}{\radian}$, $a$ and $b$ are the radius of the major and minor axis respectively, seen in Fig. \ref{fig:elliptic}. Now, the lever arm $\boldsymbol{\hat{l}_r}$ is not stationary and changes during the rotation of the knee. An example trajectory of the contact point over one full rotation of $\ang{360}$ for an ellipse with $a=3$ and $b=1$ is illustrated in Fig. \ref{fig:lever_arm}.

The contact point $\boldsymbol{\hat{l}_r}$ can be calculated using (\ref{eq:chap3:find_point}) and (\ref{eq:chap3:correct_theta_star}). Similarly, based on equations \eqref{eq:chap3:spring_torque_generic} - \eqref{eq:chap3:parameter_ellipse}, we can compute the spring displacement by numerically solving an elliptic integral. We skip the derivations for the sake of space.
The resulting torque is non-linear if $a \neq b$

\begin{equation}
    \tau_s(\boldsymbol{d}) = \psi(q_\mathrm{KFE}, \boldsymbol{d})
\end{equation}
with the design space $\boldsymbol{d} = [\bar{q}_{\mathrm{KFE}}, a, b, \phi_0]\tran \in \mathbb{R}^4$. An animation of the design space is included in the supplementary video.

\subsection{Design Optimization} \label{sec:design_opt}

\begin{figure}
    \centering
    \includegraphics[width=.43\textwidth]{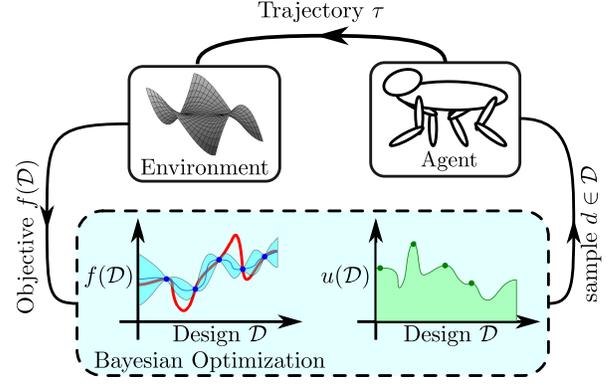}
    \caption{The trajectories $\tau$ are collected with the trained design-conditioned policy in simulation. Afterward, the robot's performance for a specific design choice is measured by a custom objective function $f(\mathcal{D})$ and sent to the \ac{bo}. Using this value, the algorithm builds a surrogate function (blue + cyan color in the left plot) and samples new points with respect to its acquisition function (green color in the right plot). The blue dots refer to already sampled points and the green dots to the next design set to be rolled out.}
    \label{fig:design_opt}
\end{figure}

Here we present our framework for optimizing the design parameters $\boldsymbol{d}$.
The general approach of our design optimization strategy is pictured in Fig. \ref{fig:design_opt}. We roll out trajectories with the design-conditioned policy (explained in section~\ref{sec:overview}) in the environment with each given set of design parameters.
We define the objective function $f$ for the design optimization by the Monte Carlo estimate over a large number of samples collected in the simulation.
By doing so, we evaluate the general performance of a design instance across many different scenarios with different initial states, disturbances, and commands.

\subsubsection{Design Objective}
The main objective of our design optimization problem is energy efficiency. Accurately simulating the efficiency of a robot is a difficult task due to various sources of energy consumption, e.g., mechanical energy at the actuators, power used to run computers and sensors, etc. We assume that the power loss of the system can be approximated by the joule heating of the individual actuators. There are other factors like transmission loss and electronics loss that are neglected. Joule heating is one of the major terms for energetic losses in electric motors and is proportional to the square of the actuator torque. Similar to the \ac{cot}, we define the \ac{cotr} as

\begin{equation}
    \mathrm{CoT} \propto \mathrm{CoTr} = \frac{\int \tau^2 dt}{mg\Delta s}
\end{equation}
with $m$ being the total mass of the robot, $g=\SI{9.81}{\meter \per \second^2}$ the gravitational acceleration and $\Delta s$ the traveled distance by the robot. By normalizing with $m$, which depends on the design, and $\Delta s$, this metric allows for the comparison of different designs and walking speeds.

\subsubsection{Optimization}

The aim of the design optimization step is to find optimal design parameters $d \in \mathcal{D}$ for a specific task $t \in \mathcal{T}$ with respect to an objective function $f(d|t, \pi): \mathcal{T} \times \mathcal{D} \rightarrow \mathbb{R}$, given the pre-trained policy $\pi$. The objective $f$ evaluates $d$ for a fixed task $t$, which is velocity tracking on rough terrain in our setup, and outputs a performance measure.

A task $t$ defines the specific problem the policy solves. These parameters could be for example terrain property (rough terrain, stairs, etc.) as well as command amplitude and direction. The task parameters are randomly sampled during policy training and design optimization.

The objective $f$ is defined by the physical quantities we are optimizing the design, e.g., joint torques or tracking performance (our setup), which are often not differentiable with respect to $d$.
In our setup, we assume $f$ is not differentiable because legged locomotion entails many discrete changes in dynamics due to foot contact. Thereby we use a black-box optimization method.



The optimization problem can then be mathematically formalized as 
\begin{align}
    \boldsymbol{d}^* &= \argmin \mathbb{E}_{\boldsymbol{d} \in \boldsymbol{\mathcal{D}}, \boldsymbol{t} \in \boldsymbol{\mathcal{T}}}[f(d | t, \pi)] \\
    \mathrm{s.t.} \quad \boldsymbol{0} &\leq \boldsymbol{c}(\boldsymbol{d}). \nonumber
\end{align}
We use the \ac{hebo} algorithm \cite{cowen2022hebo}. This \ac{bo} algorithm won the NeurIPS2020 black-box optimization challenge \cite{pmlr-v133-turner21a}. The outcome of this challenge is the reason why we chose a surrogate method over a meta-heuristic method (compare Sec. \ref{sec:robot_design}).

\subsection{Design-conditioned Policy} \label{sec:overview}
It is important to have an optimal controller for each design instance to evaluate each design instance at its best performance. We assume that we can achieve near-optimal performance with a neural network policy conditioned on design parameters. 
A recent work by Won et al. \cite{won2019learning} showed that it is possible to train a shape-conditioned policy for a bipedal robot through \ac{rl} that can maintain a stable gait while the shape of its body is dynamically changing. 

Our policy training follows the approach, and we additionally adapt the privileged learning method by Lee et. al.~\cite{lee2020learning} for sim-to-real transfer.

We train two types of policies: 
\begin{itemize}
    \item Design-conditioned policy (teacher): This policy directly observes design parameters and other environmental parameters (e.g., terrain shape and friction coefficient), which we call privileged information, from simulation. The policy is used in the design optimization loop (see Fig.~\ref{fig:design_opt}).
    
    \item Deployment policy (student): This policy is deployed on the robot with noisy measurements as observation. This policy does not have access to privileged observations and observes the history of noisy proprioceptive measurements and exteroceptive measurements. The deployment policy is explained in section~\ref{sec:student}. 
\end{itemize}
The design-conditioned policy is trained via \ac{rl} in simulation and the student policy is trained via imitation learning with simulated sensor noises. Using temporally extended observations, e.g., history of proprioceptive measurements~\cite{lee2020learning} or noisy exteroception~\cite{miki2022learning}, the student policy can estimate the privileged information and adapt to the sim-to-real discrepancy.

\begin{figure}
    \centering
    \includegraphics[width=.489\textwidth]{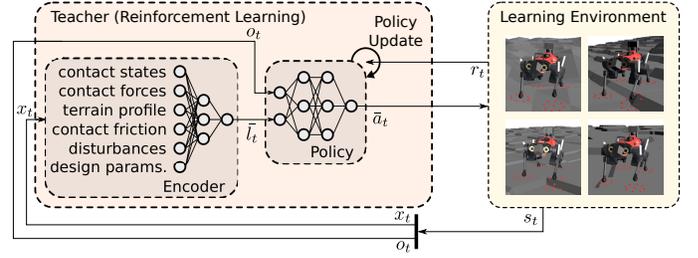}
    \caption{The learning pipeline is adapted from the teacher-student approach \cite{lee2020learning}. The most important change is that the teacher directly observes the design parameter in the privileged observation. 
    }
    \label{fig:teacher_pipeline}
\end{figure}

\begin{figure}
    \centering
    \includegraphics[width=0.489\textwidth]{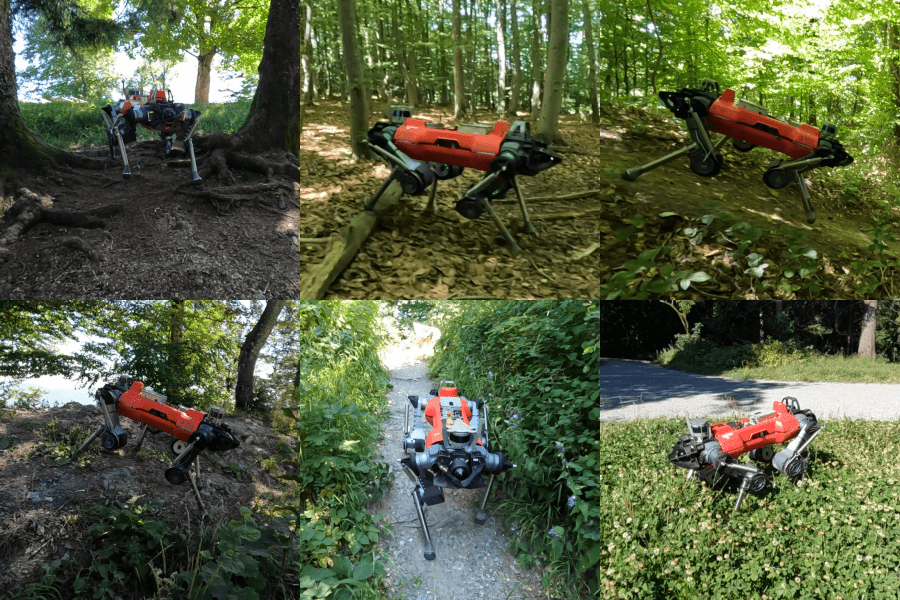}
    \caption{We performed several tests with \ac{aops} on rough terrain, showing its robustness. The policy was extensively tested in the mountains, the forests, and the City of Zurich.}
    \label{fig:hike}
\end{figure}
\subsubsection{Reinforcement Learning}
The design-conditioned policy is trained using \ac{rl}.
We model the \ac{rl} problem as a \ac{mdp}, where the design-conditioned policy $\pi_\theta$ defines the distribution of $a_t\in \mathcal{A}$ conditioned on the observation $o_t\in \mathcal{O}$. The environment updates the robots state in each step according to a transition function $p(s_{t+1} | s_t, a_t)$ and gives a reward $r_t(s_t, s_{t+1}, a_t)$. The objective is to maximize
\begin{equation}
    \pi_{\theta^*}(a_t|o_t) \rightarrow \max \mathbb{E}\left[\sum_{\tilde{t}=t}^\infty \gamma^{\tilde{t}-t} r(a_{\tilde{t}}, s_{\tilde{t}}) \right]
\end{equation}
with $\gamma \in [0, 1]$ being the discount factor.

An \ac{mdp} is defined by the 4-tuple of $ \mathcal{O}, \mathcal{A}, r, p$. The state transition ($p$) follows rigid body dynamics in simulation. Each other component is explained below.

We use the \ac{ppo} Algorithm \cite{schulman2017proximal} to update and train our policy.


$o_t$ ($\in \mathbb{R}^{133}$) contains the base target velocity commands, base orientation, base linear and angular velocity, parameters for the leg motion primitive (\ac{ftg} by \cite{lee2020learning}), a short history of joint positions and joint velocities, and the last two joint position targets. The privileged information in $\mathbb{R}^{46}$ includes contact friction, state and force at each foot, external forces and torques applied to the base, the design parameters, and the robot's link masses.

During the policy training, the design parameters are randomly sampled from $\mathcal{D}$ (compare Sec. \ref{sec:design_opt}) per episode. In order to avoid tedious design calibration, we provide observations of the equilibrium positions of the \ac{pea}s separately for each leg.

\subsubsection{Action}
The agent controls the robot through $a_t \in \mathbb{R}^{16}$ (compare Fig. \ref{fig:teacher_pipeline}), with the first 4 actions setting the frequency of the \ac{ftg} \cite{lee2020learning} and 12 additional joint position deltas. The \ac{ftg} outputs vertical foot trajectories with predefined clearance that are mapped to desired joint positions using inverse kinematics.

\subsubsection{Reward}

The reward function includes a metric for following linear base commands in the x and y directions as well as the rotation along the yaw axis. Furthermore, we punish undesired movement in the base (z velocity, roll, and pitch angular velocity). For smooth and realistic torque commands, we penalize the acceleration with which the joint position targets change over time. For the agent to find optimal and efficient behavior, we penalize the L2 norm of the actuator torques. Lastly, we penalize joint velocities that exceed the actuator limits and foot slippage, which reduces foot strain due to sliding.
\begin{figure}
    \centering
    \begin{minipage}{0.24\textwidth}
        \hfill
        \subfloat[Assembled Spring Setup
        \label{fig:assembled}]{
          \includegraphics[width=0.866\textwidth]{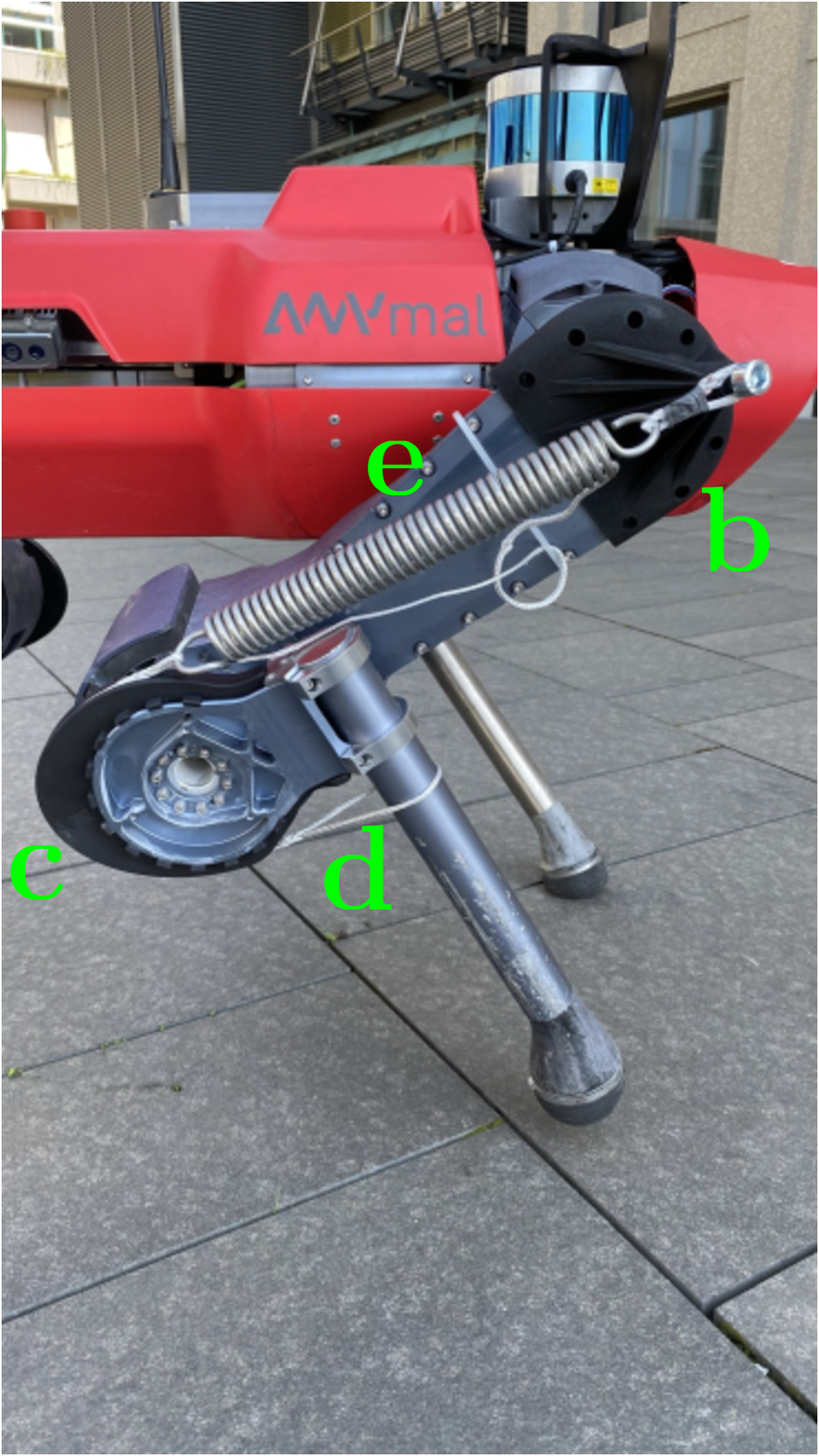}
        }
    \end{minipage}
    \begin{minipage}{.24\textwidth}
        \centering
        \subfloat[Hip
        \label{fig:hip_attachment}]{
          \includegraphics[width=.489\textwidth]{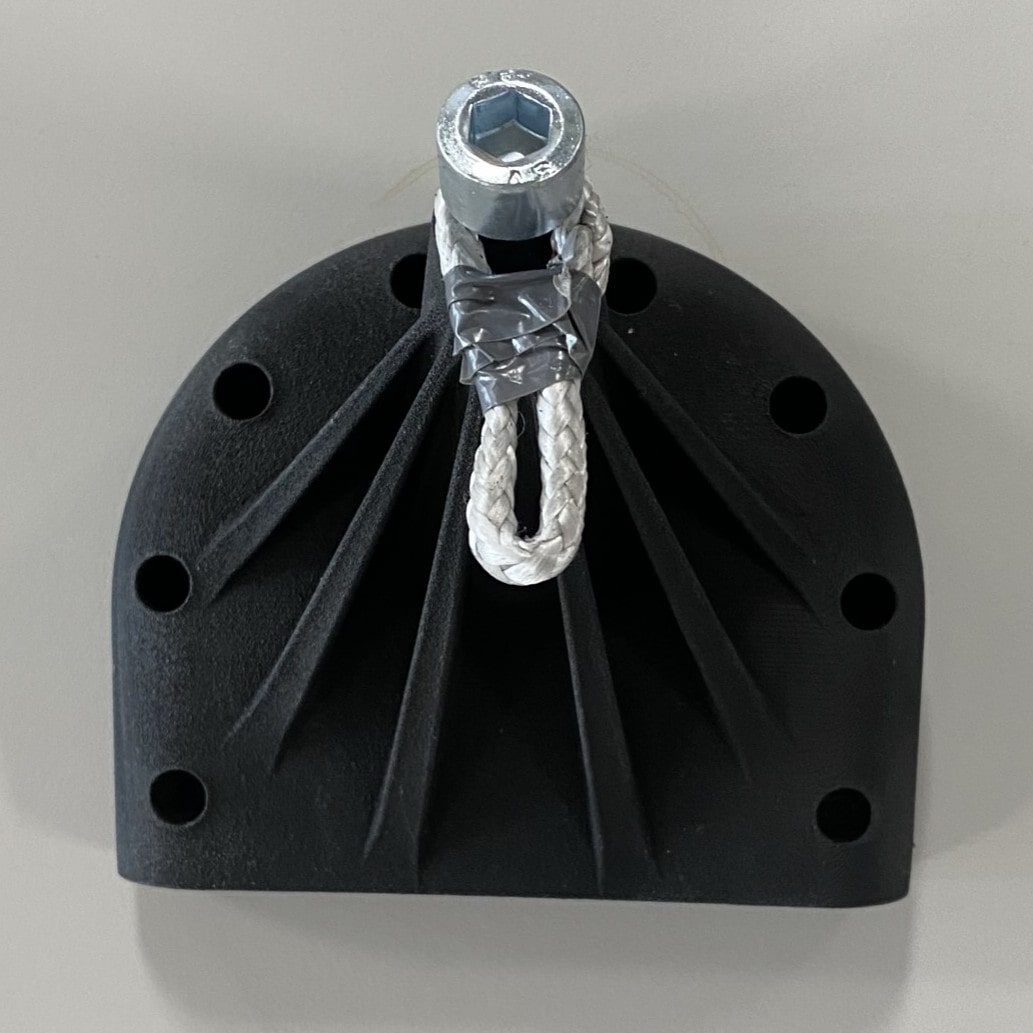}
        }
        \subfloat[Ellipse
        \label{fig:ellipse}]{
          \includegraphics[width=.489\textwidth]{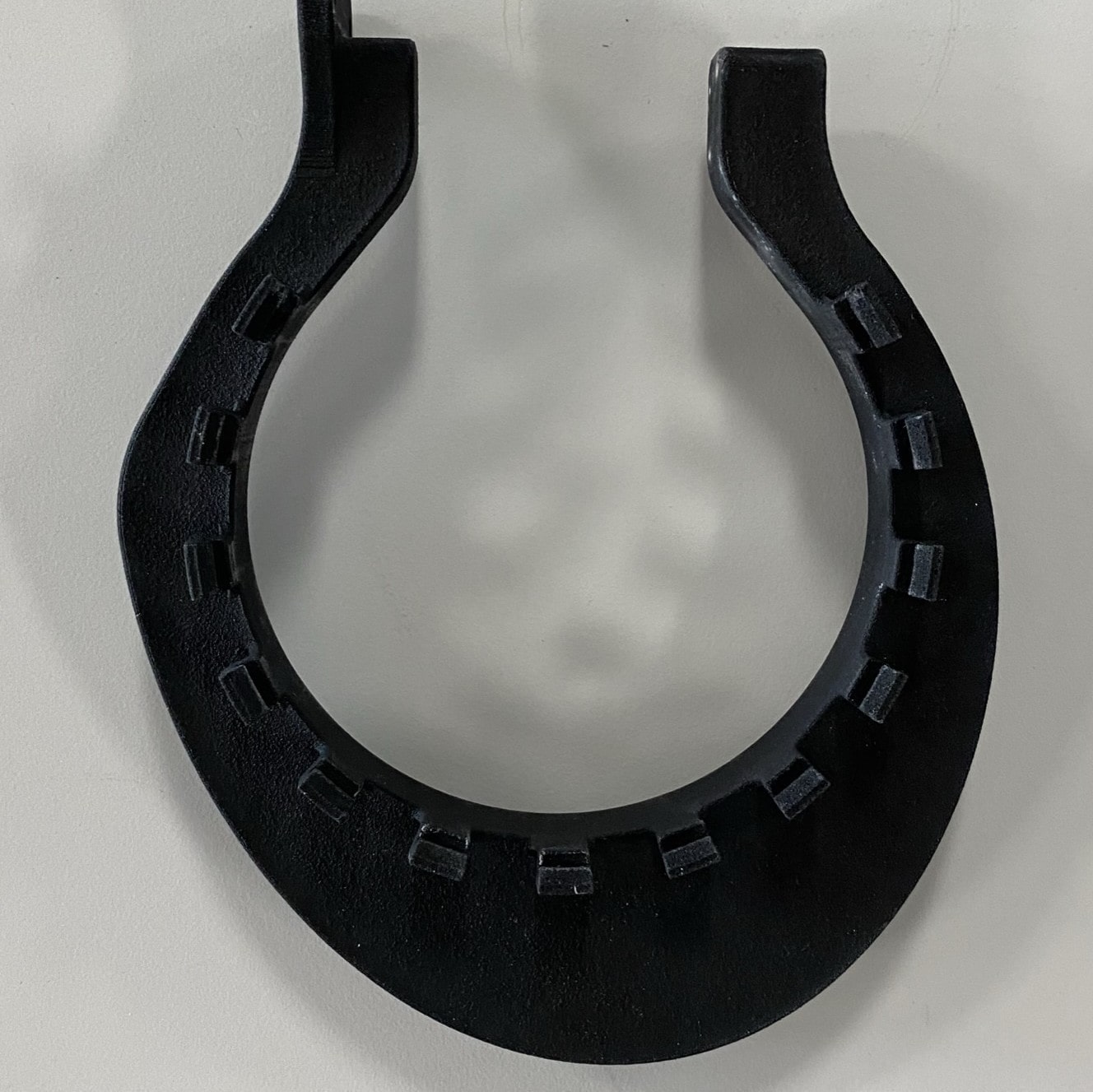}
        }
        \\
        \subfloat[Wire
        \label{fig:wire}]{
          \includegraphics[width=.489\textwidth]{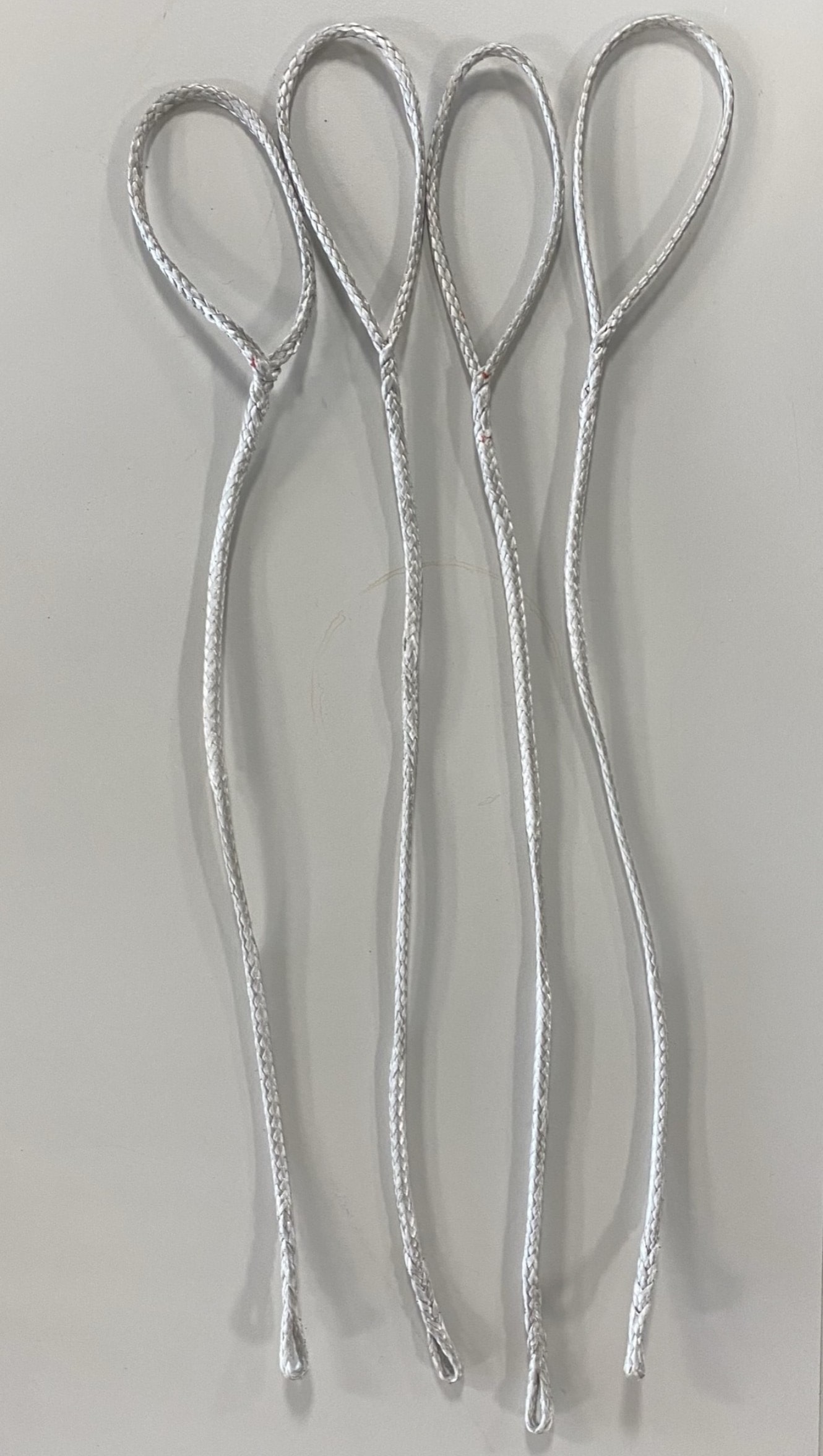}
        }
        \subfloat[Spring
        \label{fig:spring}]{
          \includegraphics[width=.489\textwidth]{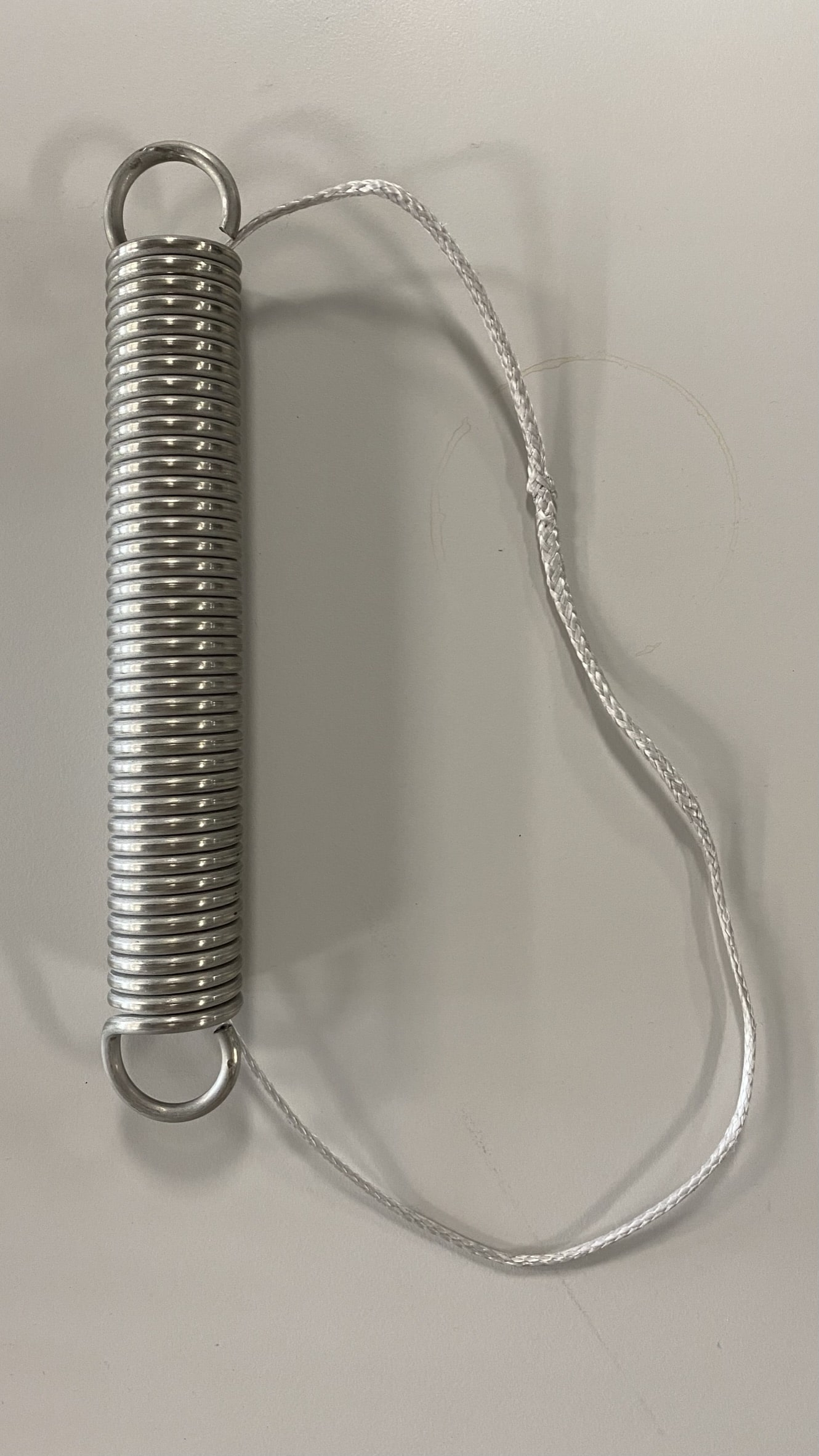}
        }
    \end{minipage}
    \caption{These images of the robot as well as the individual parts show the spring-wire setup developed in this work. The green letters in Fig. \ref{fig:assembled} correspond to the 4 pictures on the right. }
    \label{fig:hardware}
\end{figure}

\subsubsection{Architecture}
The design-conditioned policy is modeled as a \ac{mlp} and an auto-encoder network. The encoder network takes the privileged information and outputs an embedding vector $\bar{l}_t$. Finally, the proprioceptive observations and this vector $\bar{l}_t$ are used as the input to the policy network (compare Fig. \ref{fig:teacher_pipeline}). 

\subsection{Deployment Policy} \label{sec:student}

The deployed policy does not have access to privileged information. Instead, it uses a sequence of past observations to infer the unobserved state of the environment~\cite{miki2022learning}. The student policy is constructed by a \ac{rnn} \cite{cho2014learning} to effectively handle the sequential data. Similarly to Lee et al.~\cite{lee2020learning}, the training is done by imitation learning with an additional reconstruction loss for the embedding of the privileged information ($\bar{l}_t$).

The observation of the deployed policy consists of proprioceptive measurements from the IMU and joint encoders and exteroceptive measurements from depth sensors. Both modalities are simulated with noise during the training, which is not added to the design-conditioned policy's observation. The action space of the deployment policy is the same as the design-conditioned policy.

An important factor for the sim-to-real transfer is to account for the model mismatch of the springs. 
During the student policy training, the design parameters are perturbed by \SI{10}{\percent} from the optimized parameter to emulate limited manufacturing precision (see Sec. \ref{sec:setup}). The design-conditioned policy observes the exact values as privileged information while the student policy does not have direct access to the design parameter.




\section{Experiments} \label{sec:experiments}

We report the results of five different experiments to quantify the effectiveness of our approach as well as the performance gained by our new parallel elastic knee. The first experiment in Sec. \ref{sec:verify_optimization} shows that our design optimization framework can find optimal parameters with respect to our design-conditioned policy in various tasks and with high repeatability. Experiments 2 and 3, in Sec. \ref{sec:forward_walking} and Sec. \ref{sec:random_command} respectively, are hardware experiments on flat terrain, showing that the parallel-elastic robot is more efficient than the baseline and requires less torque in forward walking as well as tracking random commands. The fourth experiment in Sec. \ref{sec:hike} shows that the novel design can traverse difficult terrain. Lastly, Sec. \ref{sec:battery_life} reports the last experiment, using the robot on a running track, which shows that the newly designed robot can operate longer with the same battery charge.


\subsection{Setup} \label{sec:setup}


The task $t$ for which the robot is optimized is forward walking at \SI{1}{\meter \per \second} in an environment with stepping stones, flat terrain, and rough terrain with base perturbations of up to \SI{50}{\newton} force and \SI{50}{\newton \meter} torque. The contact friction that the robot experiences is in the range $\mu = [0.5, 2]$. The objective function $f$ is chosen as the average reward 
\begin{equation}
    f = \frac{1}{N}\sum_{i=0}^N r(a_{t_i}, s_{t_i}).
\end{equation}
We use 1000 different episodes to estimate the expectation of the objective.

We optimize the design parameters (\ref{sec:elliptic_shape}) and build the elliptic cam in Fig. \ref{fig:elliptic} for the hardware experiments. The physical parts that we created are illustrated in Fig. \ref{fig:hardware}. Our final design consists of a linear spring with stiffness $k_s=\SI{4154}{\newton \per \meter}$ and the four optimal design parameters, namely the radius of the major axis $a=\SI{8.1}{\centi \meter}$ and minor axis $b=\SI{6.0}{\centi \meter}$, initial angle $\phi_0 = \SI{0.0}{\radian}$ and the equilibrium position of $\bar{q}_{\mathrm{KFE}} = \SI{0.36}{\radian}$. The wires in Fig. \ref{fig:wire} define the equilibrium position $\bar{q}_{\mathrm{KFE}}$ of each leg and are due to manufacturing constraints not equally long. We randomize these values separately for each leg during the student training to account for unsymmetrical spring parameters. The policies use the spring exclusively in the pulling direction. Thus, we can implement the design with one tension spring per knee.


After training the design-conditioned agent, we create two student policies. For the distillation, we fix our design parameters in the demonstrations from the design-conditioned policy to the optimal design (parallel-elastic knee joint) and to $a=\SI{0}{\centi \meter}$ and $b=\SI{0}{\centi \meter}$ (rigid baseline). This allows us to create a comparative evaluation of having parallel elastically actuated knee joints with respect to the baseline. The baseline is referred to as \emph{ANYmal} and the optimal design as \emph{\ac{aops}} with a total mass of \SI{51.3}{\kilo \gram} and \SI{52.5}{\kilo \gram} respectively.

\subsection{Simulation-based Results} \label{sec:verify_optimization}
\begin{figure}
    \centering
    \includegraphics[width=.4\textwidth]{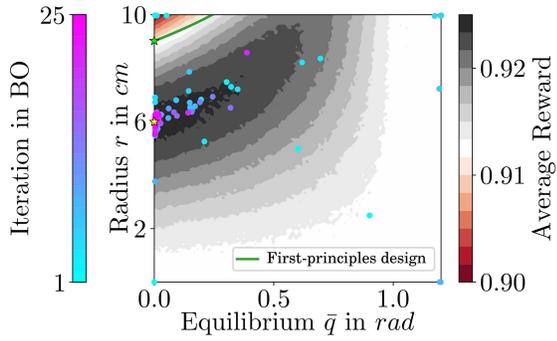}
    \caption{This figure illustrates a contour plot of the design space in the case of a linear characteristic (using a circular shape). The objective is the Average Learning Reward. Additionally, we report 25 iterations of our design optimization framework progressing from blue dots to pink dots. The green line shows the first-principles design which is derived from a conventional design approach. The yellow star indicates the optimal design and the green star is the optimal first-principles design.}
    \label{fig:contourplot}
\end{figure}
This simulation-based experiment shows that our design optimization framework can find optimal design parameters within a given interval for \ac{pea}s. In order to visualize the result, we optimize the elliptic cam from Fig. \ref{fig:elliptic} and set $a = b = r$. Therefore, since the design is point symmetric with the origin, this design has only 2 parameters $d = [\bar{q}, r]\tran \in \mathbb{R}^2$. The plot in Fig. \ref{fig:contourplot} shows a contour plot of the average learning reward in the design space $\mathcal{D}$. The contour is obtained by sampling $40$ points for each design parameter and 200 robots per design (320.000 simulated trajectories). Additionally, we report 25 iterations of our design optimization framework in Fig. \ref{fig:contourplot}. From the contour of the objective, it is observable that the optimal value lies around $\bar{q} \approx \SI{0.0}{\radian}$ and $r \approx \SI{6.0}{\centi \meter}$ (yellow star). Within the first iterations, the framework is already close to the optimal value and still explores the design space for other optimal parameters.

The green first-principles design curve in Fig. \ref{fig:contourplot} is defined by a conventional design approach. This design compensates the gravity of the robot at the average joint configuration while walking with normal ANYmal, which is \SI{1.3}{\radian}. 
We would like to minimize the torque in the flight phase ($q > 1.3$) which results in $\bar{q}$ being as small as possible. The optimal design (green star) is $\bar{q} = \SI{0.0}{\radian}$ and $r=\SI{9.02}{\centi \meter}$.

The x-axis, where $r=\SI{0}{\centi \meter}$, corresponds to our baseline since the torque is zero due to a zero lever arm. While the average reward differs in about \SI{1}{\percent}, the optimal parameter reduces the \ac{cotr} by \SI{33}{\percent} in comparison to the baseline. In contrast, the first-principles design reduces the \ac{cotr} by only \SI{8}{\percent}. This shows that our design optimization effectively finds the best design parameters given the conditioned control policy. In this case, the highest average reward results in the lowest \ac{cotr}. Please note that the \ac{cotr} is a subset of the average reward $\mathrm{CoTr} \subset \mathrm{Average Reward}$ (compare Sec. \ref{sec:overview}).

\begin{figure}
    \centering
        \subfloat[Standing
        \label{fig:tasks:standing}]{
          \includegraphics[width=0.085\textwidth]{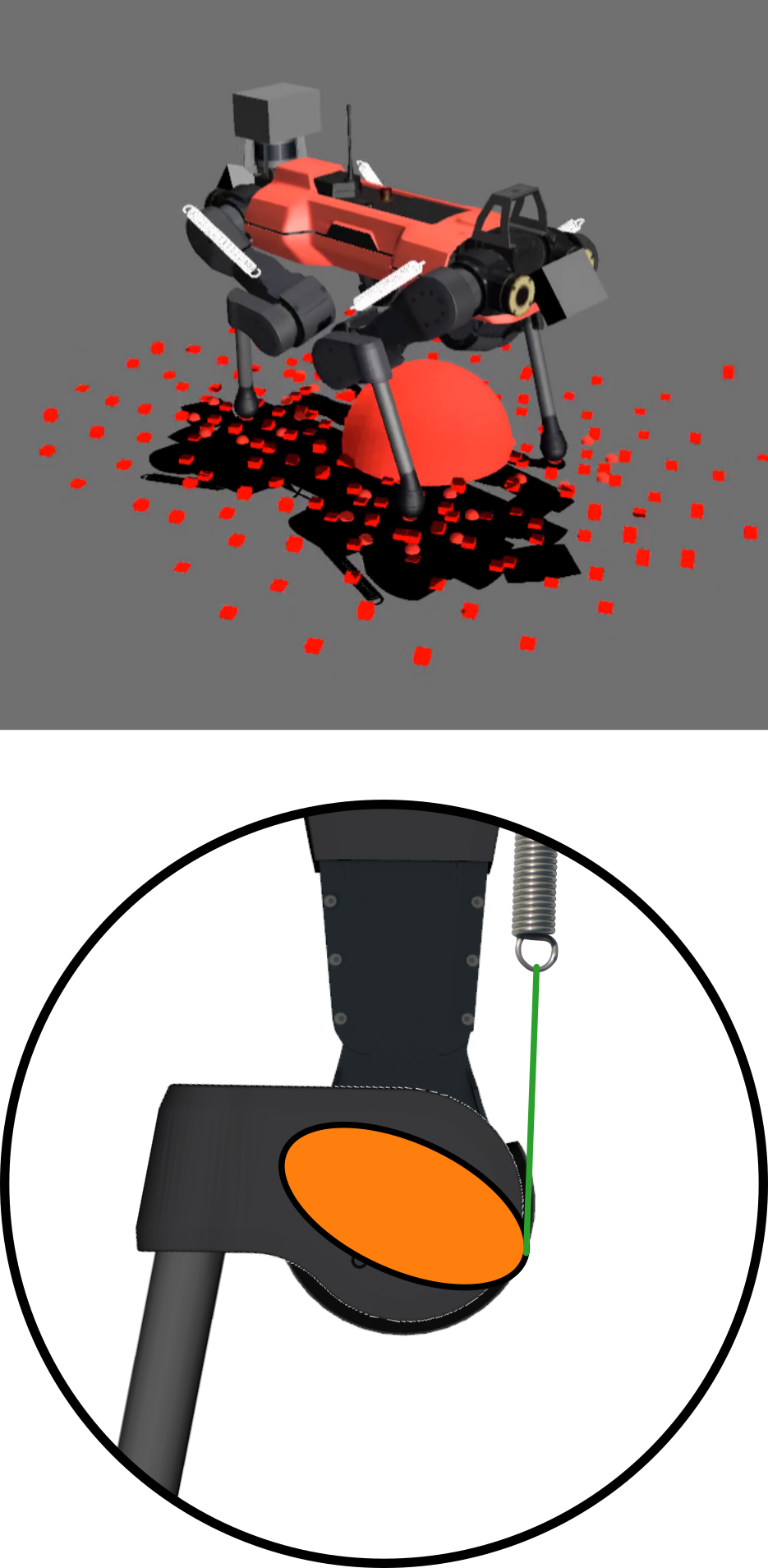}
        }
        \subfloat[Payload
        \label{fig:tasks:payload}]{
          \includegraphics[width=0.085\textwidth]{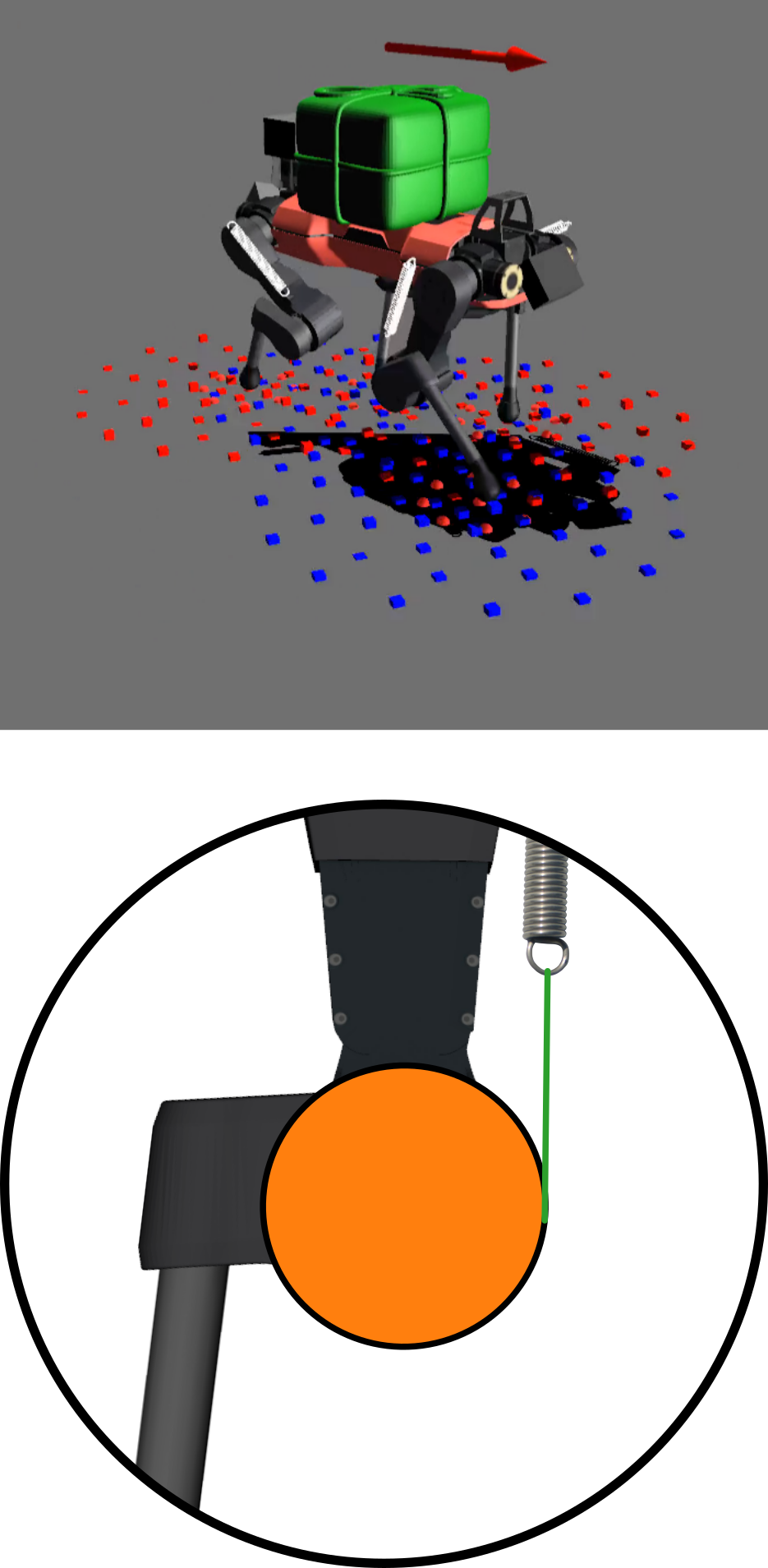}
        }
        \subfloat[Stairs
        \label{fig:tasks:stairs}]{
          \includegraphics[width=0.085\textwidth]{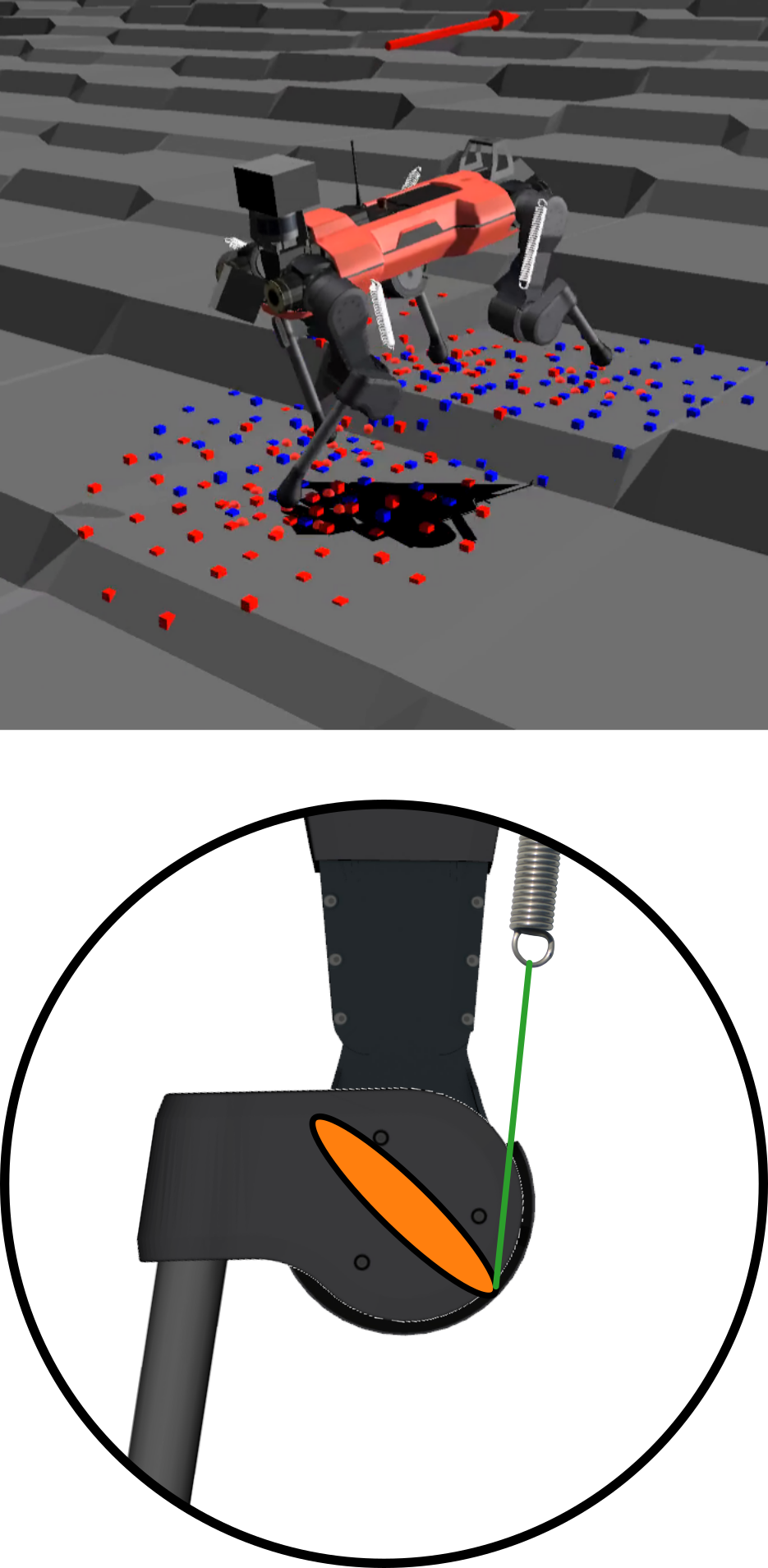}
        }
        \subfloat[Flat
        \label{fig:tasks:flat}]{
          \includegraphics[width=0.085\textwidth]{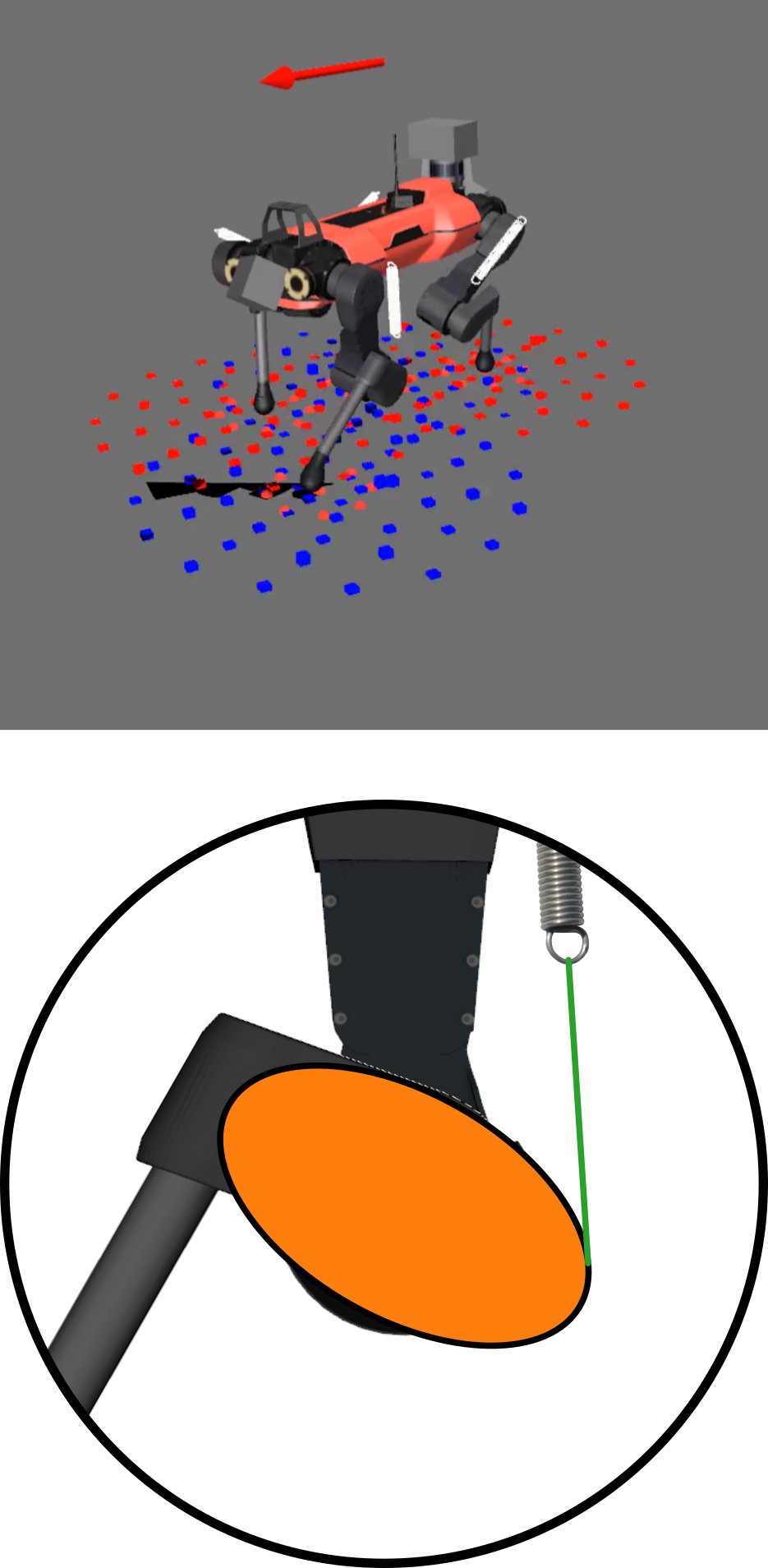}
        }
        \subfloat[Rough
        \label{fig:tasks:rough}]{
          \includegraphics[width=0.085\textwidth]{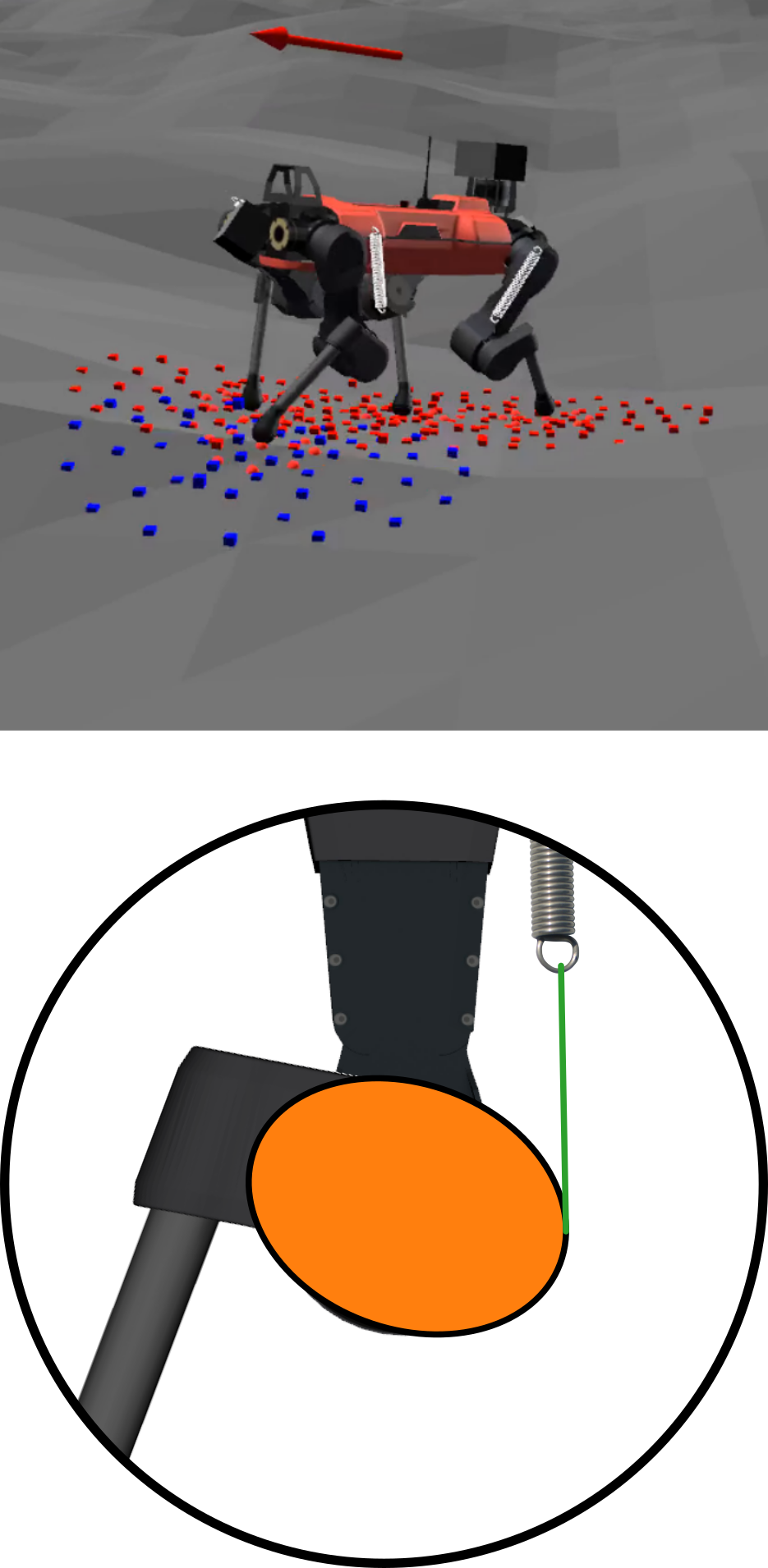}
        }
    \caption{On the top row, 5 different environments are shown for which each design is trained and optimized. From left to right, the task is standing on flat terrain, carrying \SI{20}{\kilo \gram} payload on flat terrain, walking on stairs, flat- and rough terrain. The bottom row shows each optimal design found by our framework in the equilibrium position of the spring. For the hardware experiments, we built the design in \ref{fig:tasks:rough}}
    \label{fig:tasks}
\end{figure}

Using the full design space, we trained policies with 5 different random seeds and optimized the parameters for forward walking at \SI{1}{\meter \per \second} on flat terrain (see Fig. \ref{fig:tasks:flat}). The standard deviation is below \SI{3}{^\circ} for the angles ($\bar{q}_{\mathrm{KFE}}$, $\phi_0$) and below \SI{1}{\milli \meter} for the radii ($a$, $b$).
This
shows that our design optimization method is repeatable and does not produce random designs over multiple runs. 

Finally, we optimize the design of the robot for 5 different tasks shown in Fig.~\ref{fig:tasks}
The walking experiments are optimized for \SI{1}{\meter \per \second}. The resulting designs in the bottom row show the knee configurations at the equilibrium positions and optimized cam shapes.
This result shows the effectiveness of our method for finding different optimal designs depending on different scenarios.

\begin{figure}
    \centering
    \includegraphics[width=.4\textwidth]{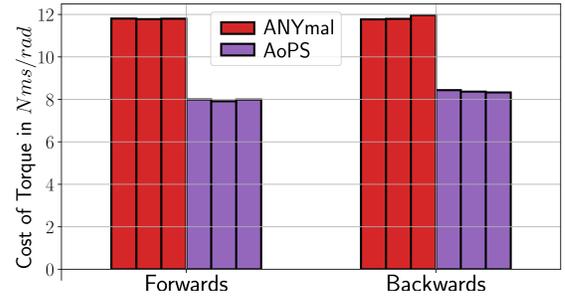}
    \caption{This bar plot illustrates the efficiency gain by adding springs on \ac{aops} (purple bars) compared to ANYmal (red bars). The former can reduce the needed torques to travel 15m by \SI{32.8}{\percent} compared to the latter.}
    \label{fig:forward_efficiency}
\end{figure}

\subsection{Forward Walking} \label{sec:forward_walking}
In this first hardware experiment, we compare the performance difference of ANYmal and \ac{aops} on flat terrain, walking \SI{16}{\meter} forwards and backward in a straight line (see supplementary video). Each robot walks $3\times$ forward and $3 \times$ backward. 

The \ac{cotr} is visualized as a bar plot in Fig. \ref{fig:forward_efficiency}. Both robots have only little variance in each test, with ANYmal experiencing a $CoTr \approx \SI{12}{\newton\meter\second}$ while \ac{aops} drives the cost down to $\SI{8}{\newton \meter \second}$. On average, \ac{aops} is \SI{33}{\percent} more efficient with respect to \ac{cotr} than the baseline ANYmal.

As shown by Fig.~\ref{fig:forward_efficiency}, our optimized design does not sacrifice the tracking performance for efficiency. Both \ac{aops} and ANYmal could track the target velocity with an error less than $0.25 \si{\meter \per \second}$.
The figure shows slightly better tracking for \ac{aops}, but the difference is negligible considering the confidence intervals (error bars).


\begin{figure}
    \centering
    \includegraphics[width=.4\textwidth]{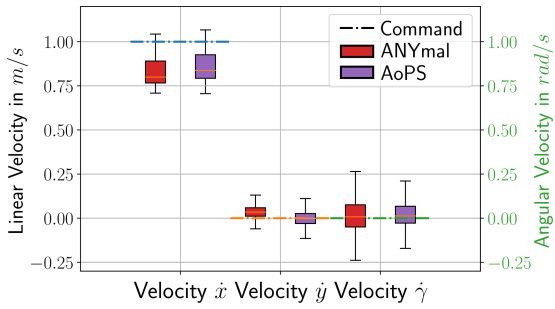}
    \caption{This figure compares the command tracking performance of ANYmal (red) and \ac{aops} (purple) for the forward walking experiment. The dashed lines show the desired velocity in the x and y direction and around the yaw axis respectively.}
    \label{fig:tracking}
\end{figure}


\subsection{Random Command Tracking} \label{sec:random_command}
In our second hardware experiment, we test how the performance translates to a more versatile task. We send 10 random commands for 3s each to the robots while the commands change dynamically (see supplementary video). The commands are randomly sampled between $[-1.2, 1.2] \si{\meter \per \second}$ in x direction, $[-0.6, 0.6] \si{\meter \per \second}$ in the y direction, and $[-1.2, 1.2] \si{\radian \per \second}$ around the yaw axis and the same for both robots. The efficiency gain for the execution of all the commands is again around \SI{30}{\percent} for \ac{aops} while the tracking performance was similar to ANYmal.

\begin{figure}
        \centering
        \subfloat[ANYmal
        \label{fig:torque_anymal}]{
          \includegraphics[width=.23\textwidth]{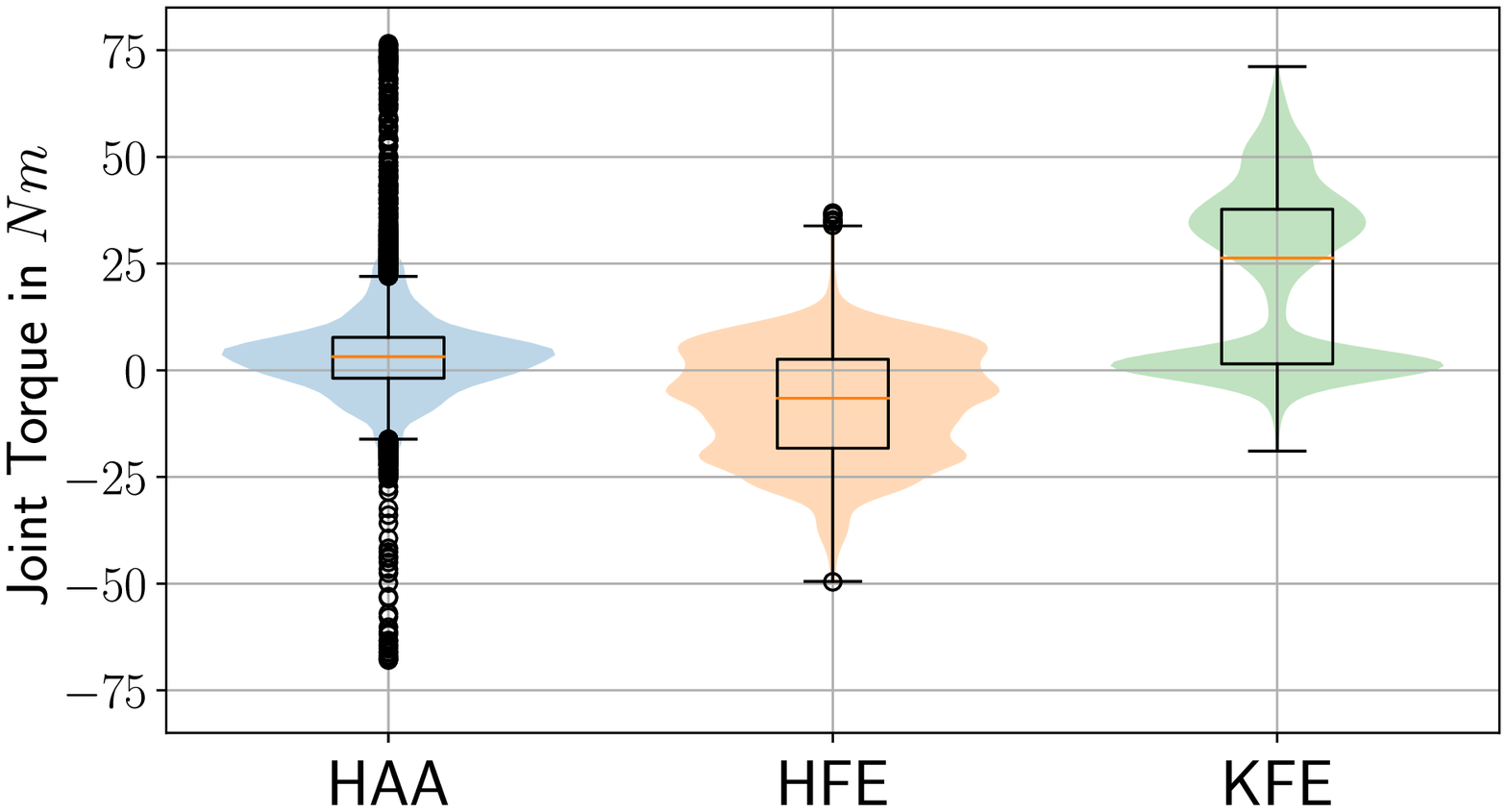}
        }
        \subfloat[\ac{aops}
        \label{fig:torque_aops}]{
          \includegraphics[width=.23\textwidth]{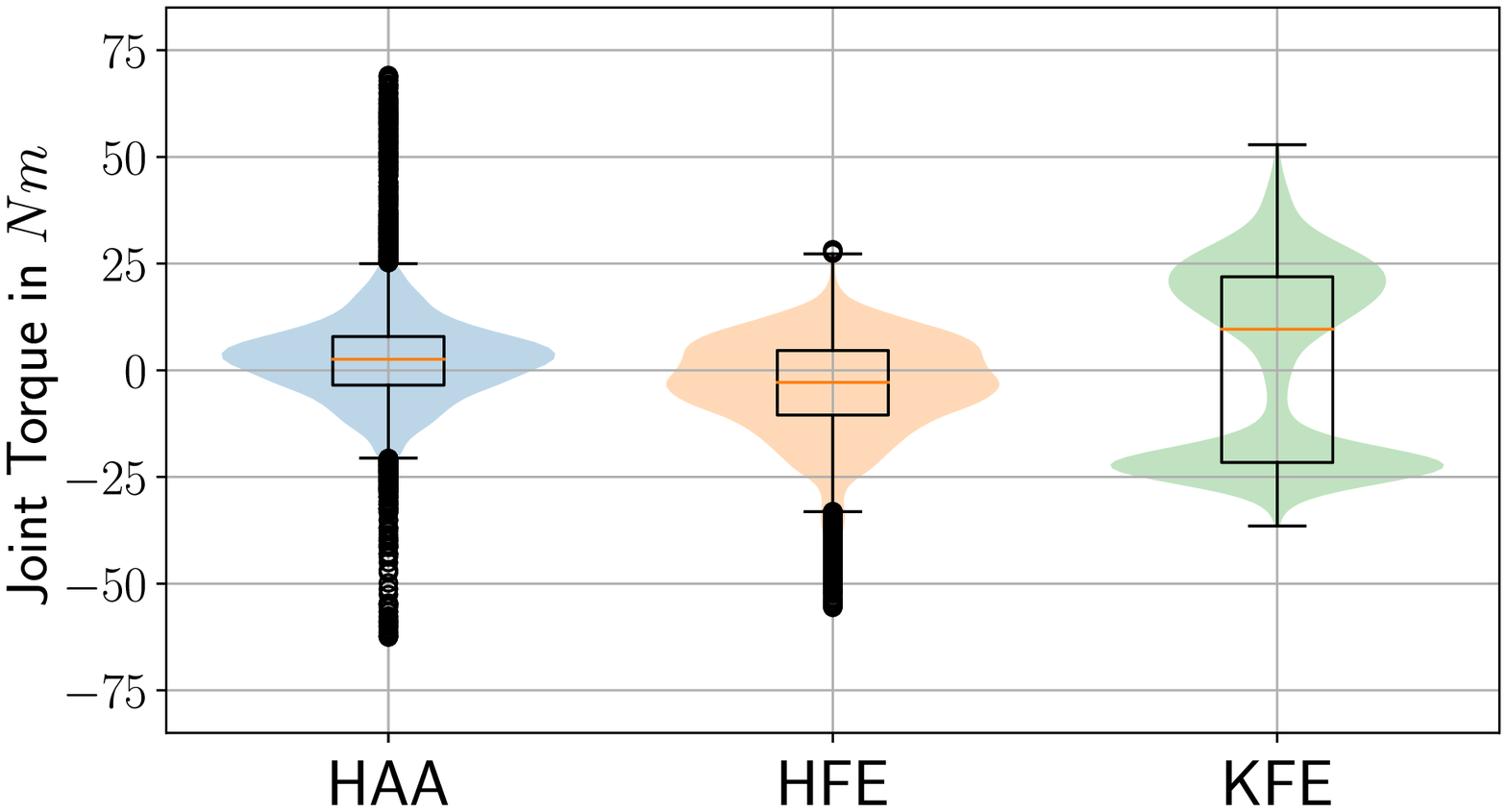}
        }
    \caption{These two graphs show box plots of the torques needed for each joint separately in one experiment where the robots are tracking the 10 random commands. For readability reasons, only the \ac{LF} leg is presented. Furthermore, the distribution for each joint torque is indicated by colored violin plots. This plot transfers similarly to the other legs as well.}
    \label{fig:torque_boxplot}
\end{figure}

Additionally, Fig. \ref{fig:torque_boxplot} reports the joint torques for the left front leg of the robots as a boxplot with an overlaying violin plot. Regarding the KFE joint (knee), the average torque is around \SI{26}{\newton \meter} for ANYmal in Fig. \ref{fig:torque_anymal} while \ac{aops} is around \SI{7}{\newton \meter}. Basically, the whole distribution shifts down thanks to the parallel elastic spring, which reduces the \ac{cotr} notably. As a result, the maximum absolute torque that \ac{aops} needs for the same task is \SI{52}{\newton \meter}, which is only \SI{71}{\percent} of ANYmal (\SI{73}{\newton \meter}). Furthermore, the HFE joint average torque for \ac{aops} is closer to \SI{0}{\newton \meter} than ANYmal, while at the same time requiring less variance. This also drives down the \ac{cotr}. Expectedly, the HAA joint is unaffected by the parallel elastic spring, and for both systems mostly the same.

\subsection{Rough Terrain} \label{sec:hike}
For the fourth and fifth tests, we adapted the perceptive learning from Miki et. al. \cite{miki2022learning} and included exteroceptive observations during the student distillation. Using this adapted policy, we performed several outdoor experiments with our parallel-elastic robot. We climbed several inclinations, traversed different types of stairs, went through confined spaces, walked over forest ground, inclined gravel paths, etc. A few snapshots are presented in Fig. \ref{fig:hike} and videos in the supplementary material. The robot did not fall once during the tests and reports the robustness of the controller and the novel design. This shows that adding parallel elastic springs does not affect the robustness negatively.

\begin{figure}
    \centering
    \includegraphics[width=0.4\textwidth]{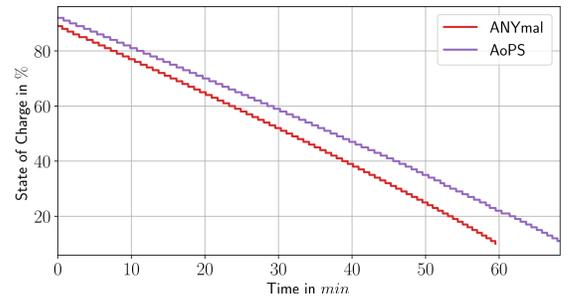}
    \caption{The state of charge for ANYmal (Red) and \ac{aops} (Purple) over time during the experiment in Sec. \ref{sec:hike} shows that our optimized design can achieve higher operating times with the same battery.}
    \label{fig:state_of_charge}
\end{figure}

\subsection{Battery Life} \label{sec:battery_life}
\begin{table}
\centering
\caption{\small Running track performance}
\label{tab:energy_efficiency}
\begin{tabular}{c|c|c} 
  & \ac{aops} & ANYmal \\
 \hline
 Number of Rounds & 7.5 & 6.6  \\ 
 \hline
 Traveled distance [\si{\meter}] & 3000 & 2640  \\ 
 \hline
 Initial Charge [\si{\percent}] & 92 & 89  \\ 
  \hline
 Final Charge [\si{\percent}] & 11 & 10 \\ 
 \hline
 Operation Time [\si{\minute}] & 68 & 59 \\
 \hline
 Average Velocity [$\nicefrac{m}{s}$] & 0.735 & 0.740 \\
 \hline
 Efficiency [\si{\percent}] & 111 & 100 \\
 \hline
 Outside Temperature [\si{\celsius}] & 31 & 26
\end{tabular}
\end{table}

Finally, we used both robots sequentially on a running track of \SI{400}{\meter} length and let the robots walk with the same battery until the battery was fully depleted. The battery was as much as possible fully charged before and after the first run with \ac{aops} to ensure a fair evaluation. Both robots were commanded \SI{1}{\meter \per \second} and carefully steered to stay in the inner path of the track. The performance of each robot is reported in Tab. \ref{tab:energy_efficiency}.  This experiment shows that the overall traveled distance of our quadrupedal robot can be increased by at least \SI{11}{\percent} from \SI{2640}{\meter} to \SI{3000}{\meter}. We introduce the following efficiency metric as the quotient in covered distance scaled by the mismatch in battery charge (\SI{2}{\percent}).
\begin{equation} \label{eq:efficiency}
    \mathrm{Efficiency} = \frac{\SI{3000}{\meter}}{\SI{2640}{\meter}} * \frac{0.89 - 0.10}{0.92-0.11} = 1.11.
\end{equation}
We also report the state of the charge over time in Fig. \ref{fig:state_of_charge}. Besides the faster drop for ANYmal, this shows that the battery that we used is internally calibrated and the linear scaling in \eqref{eq:efficiency} can compensate for the \SI{2}{\percent} difference in charge.

\section{Conclusion}
This paper shows that, with the co-optimization of the design and controller, parallel springs on the knee of quadrupedal robots can increase locomotion efficiency without compromising the command tracking performance and robustness.
While it is well studied that gravity compensation with \ac{pea}s is energetically beneficial for static tasks \cite{kashiri2018overview}, the \ac{pea}'s contribution during the dynamic locomotion is relatively unstudied. The effect of \ac{pea} is nontrivial during the locomotion since the actuators have to repeatedly work against the spring. 
A key takeaway of our work is that \ac{pea}s can also increase the performance during dynamic locomotion.

We co-optimized design parameters and locomotion controllers that act optimally for a given set of design parameters and task. With a parallel elastic knee actuator designed by our approach, we could reduce the required joint torques, which yields a higher operation time for our quadrupedal robot ANYmal during locomotion.


An important thing to note from our hardware experiments is the robustness of our controller to the model uncertainty, which shows the practical benefit of the \ac{rl}-based control method. Trained by the privileged learning method~\cite{lee2020learning} with randomized spring parameters, our controller tolerates possible model mismatches on the physical system without accurate spring calibration procedures, thus, removing the need to run any complex system identification routine.

As we showed the potential of \ac{pea}s in legged robotics, further investigations in this direction have to follow.
Firstly, the physical system's energy consumption must be better modeled.
This work assumes that the \ac{cotr} measurement is proportional to the battery life of the robot. Nevertheless, during the experiments in Sec. \ref{sec:forward_walking} and Sec. \ref{sec:battery_life}, we found a discrepancy. 
There are unmodeled factors such as electrical and mechanical losses which we did not identify in this work.
Secondly, the design-conditioned policy cannot be guaranteed to be as performant as a policy trained for each design parameter. The discrepancy was negligible in the setup covered in this paper. A previous study on this topic was conducted by us~\cite{belmonte2022meta}.
Lastly, a more elaborate design should be introduced. Our current design limits the workspace of the knee joint and the implementation of the cable-spring mechanism can be inaccurate. Additionally, research will be devoted to including other parameters in the design process like link masses or leg lengths.
\section*{Acknowledgment}
The authors would like to thank the RSL Design Team for their insightful discussions and Marko Bjelonic for his great support on the Cluster and for helping with the state estimation on \ac{aops}.

\ifCLASSOPTIONcaptionsoff
  \newpage
\fi



\bibliographystyle{IEEEtran}
\bibliography{IEEEabrv,references}
%

%








\end{document}